\documentclass[preprint,10pt,authoryear,5p,twocolumn]{elsarticle}

\usepackage{amssymb}

\usepackage{natbib}
\journal{Digital Forensics Research Conference Europe 2025}

\usepackage{caption}
\usepackage{subcaption}
\usepackage{float}

\usepackage[]{todonotes}

\usepackage{xspace}
\newcommand{\ie}{i.e.\@\xspace} 
\newcommand{\eg}{e.g.\@\xspace} 

\usepackage{url}

\begin{document}

\begin{frontmatter}



\title{Exploring the Robustness of AI-Driven Tools in Digital Forensics: A Preliminary Study}


\author{Silvia Lucia Sanna, Leonardo Regano, Davide Maiorca, Giorgio Giacinto}

\affiliation{organization={Department of Electrical and Electronic Engineering},
            addressline={University of Cagliari}, 
            city={via Marengo 2, 09123, Cagliari},
            country={Italy}}

\begin{abstract}
Nowadays, many tools are used to facilitate forensic tasks about data extraction and data analysis. In particular, some tools leverage Artificial Intelligence (AI) to automatically label examined data into specific categories (\ie, drugs, weapons, nudity). However, this raises a serious concern about the robustness of the employed AI algorithms against adversarial attacks. Indeed, some people may need to hide specific data to AI-based digital forensics tools, thus manipulating the content so that the AI system does not recognize the offensive/prohibited content and marks it at as suspicious to the analyst. This could be seen as an anti-forensics attack scenario. 
For this reason, we analyzed two of the most important forensics tools employing AI for data classification: Magnet AI, used by Magnet Axiom, and Excire Photo AI, used by X-Ways Forensics. We made preliminary tests using about $200$ images, other $100$ sent in $3$ chats about pornography and teenage nudity, drugs and weapons to understand how the tools label them. Moreover, we loaded some deepfake images (images generated by AI forging real ones) of some actors to understand if they would be classified in the same category as the original images. From our preliminary study, we saw that the AI algorithm is not robust enough, as we expected since these topics are still open research problems. For example, some sexual images were not categorized as nudity, and some deepfakes were categorized as the same real person, while the human eye can see the clear nudity image or catch the difference between the deepfakes. Building on these results and other state-of-the-art works, we provide some suggestions for improving how digital forensics analysis tool leverage AI and their robustness against adversarial attacks or different scenarios than the trained one.
\end{abstract}




\end{frontmatter}

\section*{Disclaimer}
Content Warning: This document contains content that some may find disturbing or oﬀensive,
including content that is sexual, hateful, or violent in nature.
\section{Introduction}
\label{sec:intro}
Digital Forensics (DF) includes a set of different techniques to retrieve data from every digital device. The data extraction and analysis can be performed even if the device is turned on (live forensics with non-repeatable steps) or off (post-mortem analysis with repeatable analysis). This science is applied to understand what happened when a cybercrime is committed, but also when a traditional crime is perpetrated, and a digital device can contain important information to solve the case. The forensic analyst must follow specific standards to manage the evidence; otherwise, it may not be admissible in legal proceedings. Moreover, the analysis should be performed using suitable techniques according to the type of device to be analyzed and also the data of interest found in it. For example, an Android smartphone is analyzed differently than a Windows computer, or a jpg image differently than a png image. This is because of the singular and unique structure of each device, OS, and file format. 

Different tools have been developed to help humans analyze different devices and data. Indeed, interpreting data in a manual, bit-per-bit fashion is extremely complex and time-consuming for a human analyst. Thus, many more tools have been developed for disk analysis: Autopsy\footnote{\url{https://www.autopsy.com/}}, FTK Imager\footnote{\url{https://www.exterro.com/digital-forensics-software/ftk-imager}}, Magnet Axiom\footnote{\url{https://www.magnetforensics.com}} Processing and Examine, and X-Ways Forensics\footnote{\url{https://www.x-ways.net/forensics/}}. Traditional tools rely on known pattern recognition (\ie, predefined rules, signatures, magic number for file identification, metadata analysis, hash matches for malicious file identification, keywords search, etc.) to extract files from the raw hexadecimal file, which is the dump of the memory of the seized evidence. These tools use deterministic methods and require significant human input at each investigation stage with a manual process to label and analyze the items in the evidence. The traditional analysis is time-consuming for the analysts who have to manually check every result, and it gets more complicated when a massive amount of data must be analyzed (traditional tools can struggle to process everything efficiently). Moreover, as tools rely on database matching to check the presence of malicious files, newer malware cannot be detected. 
Recently, to ease the analyst burden, some tools (\eg Magnet AI) have been integrated with Artificial Intelligence (AI) algorithms to automatically recognize and label specific files found in the device. These tools identify patterns, anomalies, and relationships that would be difficult or time-consuming to detect manually. Indeed, Machine Learning (ML) algorithms may be employed to detect the content of specific images and label them in the proper category (\eg adult content, violence, weapons, drug, legal or personal files), while Natural Language Processing (NLP) techniques may be useful to understand the content of emails, chats and documents (\eg threatening, financial fraud, luring, etc.). Other tools such as X-Ways Forensics perform facial recognition on images and videos to match against known individuals or other faces in the acquired evidence. In summary, we can say that different applications of ML and computer vision have been integrated into Digital Forensics tools to help the analyst label the pictures according to the legal practitioners needs. This could be very helpful when for example the analyst must inspect the device of someone accused of child pornography, and such device contains many non-relevant pictures, which may be filtered through an AI-based automatic analysis. Indeed, specific files of interest (according to the subject of legal accusation) may be dumped without analyzing all the evidence, thus avoiding the need to inspect non-interesting or too private data. Moreover, it is useful to prevent stress and shock in forensic analysts when, for example, child pornography is prosecuted. Many analysts suffer psychological damage after seeing specific pictures \citep{Sanchez19_DI}.

While the benefit of employing AI in this scenario is evident, the potential impact of adversarial attacks must be taken into account. Indeed, multiple attack techniques able to mislead AI algorithms in making autonomous decisions have been reported in the literature \citep{Carlini17_ISSP, Papernot17_AsiaCCS, Biggio13_ECML}. 
The attacker's goal is to find a slight perturbation of an input that is able to trigger an erroneous classification from the targeted AI algorithm (\ie the perturbed input is classified with a different label than the original input). In particular, adversarial attacks in computer vision mean finding a perturbation (not noticeable by the human eye) that, when applied to a picture, makes the AI algorithm recognize it as another class. 
In the DF scenario, such attacks may consist in perturbing images on a sensitive device, so that, if the latter is subsequently included as evidence in a legal case, an analyst may be hindered from retrieving relevant images using an AI algorithm. The technique can also be applied to alter the audio of a target person in order to make it similar to another person's voice (\eg a singer, a politician). The rationale of this attack may be fabricating multimedia content necessary to prove that a person said sentences while, in reality, he or she did not. Such techniques are collectively known with the umbrella term \textit{deepfake}. 

Thus, we wondered if the same concept could be applied to the AI-based black-box algorithms used to automate Digital Forensic tools. We claim that cybercriminals or expert users can also use a proper perturbation to hide data on their devices, making the analysis more complicated (anti-forensic technique). 

This paper will aim to address the following research questions: \emph{(i)} \textbf{How robust are those tools and what is the false positive detection rate?} This means how many of them would be wrongly classified by giving a set of pictures belonging to the same class but different from the standard view or scenario (\ie using pornographic comic pictures; pictures of dresses with nude drawn on it; interracial pictures). \emph{(ii)} \textbf{What happens if we input adversarial images such as AI-generated pictures (deepfake) of faces similar to known people for face recognition?} This test is made to quantify the tool's robustness in recognizing the same subject in a fake context such as those generated by deepfake. \emph{(iii)} \textbf{Could adversarial attacks be used by criminals for anti-forensics purposes?} By applying these techniques, some people could perturb their pictures so that the human eye can still see them (\eg, pornography pictures, fake pictures of the same people), but the AI algorithm does not and, in this way, perturb the result of a fully automated forensics analysis.
In summary, our main contribution is the robust evaluation of the most popular digital forensics tools using black box AI to improve the analysis and detection of specific evidence and if there is a way to evade and mislead such systems. 

The remainder of this paper is structured as follows. \textit{Section~\ref{sec:foren-tools}} presents a technical background about the most popular forensic tools divided by the nature of the analyzed data, with details on the AI tools. Previous research on this topic is illustrated in \textit{Section~\ref{sec:sota}}, while \textit{Section~\ref{sec:experiments}} presents the conducted analysis, the created dataset and the methodology with research questions for each AI-based forensic tool examined. Experimental results are reported in \textit{Section~\ref{sec:results}} and discussed in \textit{Section~\ref{sec:discussion}} with the summary answers to the presented research questions.  Finally, \textit{Section ~\ref{sec:future_work}} discusses current research limitations and the future works that may improve this work and, in general, the use of AI in Digital Forensics tools.

\section{Background: Digital Forensics Tools}
\label{sec:foren-tools}
Before analyzing any type of forensic evidence (\eg documents, pictures, etc.), we need to acquire the digital evidence while preserving its integrity. The physical acquisition allows to analyze the device without altering the evidence and creating an exact bit-by-bit copy of the entire device at acquisition time to make it repeatable and reproducible, capturing all data sectors, including areas that may not be visible or accessible through the file system (\eg deleted data, unallocated space, hidden partitions). Instead, the logical acquisition is a simple copy of all the files, including directories, subdirectories and databases or other structured data that can be easily accessed by the operating system.
However, in a live analysis, results may vary depending on the system state (\eg RAM contents) at acquisition time. Thus, repeated acquisitions may lead to different results. The acquisition can be performed using different tools according to the device type and object. 
For disk acquisition, we can use \textit{FTK Imager} for Windows and the \textit{dd} command in Linux to acquire a disk with a Linux partition or Android devices.

After the acquisition, 
the analysis typically starts on the copy of the acquired copy of the evidence. To preserve the integrity, we should always make two copies so that if one is broken, the other one can be used as a backup. 
The second main step is the analysis, which uses different tools according to the data type. 
FTK Imager can be used to analyze any disk running Windows, Linux, mobile devices. 
\subsection{Digital Forensics Tools with Artificial Intelligence}
\label{sec:foren_tools:subsec:tools_ai}
As described in the Section~\ref{sec:intro}, AI algorithms have been integrated to DF tools to improve the large scale data analysis by recognizing specific data of interest and reducing the stress produced by the manual analysis of the forensics analyst.
To the best of our knowledge, only two tools use Artificial Intelligence to improve their analysis while writing this paper. Magnet Axiom uses Magnet AI to identify chats containing or suggesting grooming/luring content or sexual conversation and to look for pictures for evidence of drugs, weapons, nudity, and child abuse material. The other one is X-Ways Forensics, which uses Excire Photo AI (also available separately, \footnote{\url{https://excire.com/en/excire-search/}} 
to detect photo content automatically, find similar photos, and identify faces of known relevant people in photos by simply typing their full name or even their surname. It should be noted that they published the list of known objects but not the list of known people. As declared by both websites, no data related to the case under analysis is loaded in any cloud to be labeled. On the contrary, while installing their product, the pre-trained AI algorithm will be installed and run locally on the used machine for the analysis. This is important to preserve privacy. Instead, no information is released about the AI algorithm used to train the system; we only know it is proprietary. 

\section{Related Works}
\label{sec:sota}

Different works in the literature tried to address the problem. In particular, a study of $2018$ analyses the robustness of some DNNs used in DF \citep{Aditya18_TrustCom}. The authors propose an "Adversary Testing Framework (ATF)" to generate adversarial attacks that can bypass black-box DNNs defenses in constrained environments with fewer queries and minimal perturbations. The authors applied the ATF to the commercial tool \textit{Image Analyzer} (the version cited in the paper is no longer available): it is a DNN-based forensic tool able to flag offensive content in images (e.g., terrorism, weapons, and pornography). The system's detection is evaded with high confidence using crafted adversarial examples. The authors also propose new metrics such as attack success rate, average distortion, and query limitation to evaluate the performance of adversarial attacks. We could not compare our results with that study because the used API is no longer available, and over the years, several tools called "Image Analyzer" have been released. Adversarial attacks on image forensics have also been addressed in \citep{Nowroozi20_Siena} by developing techniques to improve ML robustness. The detected manipulations are double JPEG compression and contrast adjustment, and they also suggest improving robustness with adversarial training, as done by other works. According to this line, we tested two of the most used tools of DF using AI (Magnet and X-Ways Forensics) to understand their robustness, their behavior with adversarial samples and define how to improve them.

Other papers analyzed the use and impact of Artificial Intelligence in Digital Forensics. First of all, we need to distinguish between the use of AI in Digital Forensics tasks and AI forensics: the first one is the application of AI algorithms for digital forensics purposes, while AI forensics \citep{baggili19_arXiv} is the investigation of AI algorithms outputs.

\subsection{Artificial Intelligence for Digital Forensics}
\label{sec:sota:subsec:aidf}
Most of the \textit{AI applications to DF} are for deepfake detection, age estimation, and image classification, as highlighted by \citep{Schneider23_JISA}. In the paper, the authors review key subfields of DF that leverage AI, such as multimedia analysis (e.g., deepfakes and forgery detection), and outline the current challenges, including handling and processing vast datasets, scaling AI models to manage increasing forensic data without compromising accuracy and efficiency, the need for updated legal frameworks and ensuring algorithm robustness against adversarial attacks. They also highlight future directions, focusing on improving automation and efficiency, enhancing AI robustness through adversarial training, and addressing issues with noisy, incomplete, or low-quality data. Furthermore, they stress the importance of cross-validation and explainable AI (xAI) \citep{Gunning19_ACM, Ribeiro16_ACM} to ensure transparency in how algorithms associate data with specific labels. The xAI in DF has also been addressed by \citep{Solanke22_FSIDI}, highlighting the problem of black-box AI algorithms in DF tools, which provide results without a clear, understandable explanation of how the label has been assigned. According to the authors, the AI algorithms in DF tools should be both explainable (with a focus on the legal importance), interpretable, and transparent, hence proposed an AI framework with such characteristics. The use of AI in DF is also described by \citep{Dunsin24_FSIDI}, where AI and ML can be used for data collection, processing, and analysis. The authors claim that pattern recognition, rule-based reasoning, and genetic algorithms are explored to optimize forensic processes, enabling analysts to detect key evidence faster and more accurately, especially in areas like image analysis, malware detection, and behavioral analysis. As with the other works, they also highlight the importance of using xAI to better understand the classification, even from a legal perspective.

\subsection{Deepfake detection}
\label{sec:sota:subsec:deepfake}
Regarding the topic of our analysis, several papers have been published for \textit{deepfake detection}. In our work, we also used this topic as a test case to evaluate the robustness of Excire Photo AI in identifying known relevant people's faces in photos. A deepfake is an AI-generated digital video, audio, or image that is highly realistic but with fake content. Typically, it is used for swapping faces, mimicking voices, and creating new false scenarios but making them appear real. Deepfakes have been used for entertainment, malicious activities, misinformation, fraud, and identity deception. Different techniques have been developed to detect the different categories of deepfakes \citep{Concas22_mdpi, Gao24_EAAI}, for example in the video recognizing the eye blinking \citep{Li18_WIFS}, the inconsistent head poses \citep{Yang19_ICASSP}, the perceived emotions \citep{Mittal20_ACM} or using Recurrent Neural Networks (RNN) \citep{Guera18_CAVSBS} and Convolutional Neural Networks (CNN) \citep{Amerini19_ICCVW}. Many techniques have been investigated for face manipulation as well examined by \citep{Tolosana20_IF} or brand new techniques as the one proposed in \citep{Panzino24_CVPR}. Also the reconstruction of original images from deepfakes have been analysed such as in \citep{Guarnera22_MDPI} . Other techniques have been addressed for audio deepfakes as in \citep{Hamza22_IEEEAccess, Yi22_ICASSP, Almutairi22_scopus}. As we can see from the literature, deepfake creation, manipulation, reconstruction and detection is still an open research work.

\subsection{Nude detection}
\label{sec:sota:subsec:nude}
The second set of experiments is focused on \textit{nudity, pornography, and child pornography identification}. As highlighted by \citep{AlDahoul21_mdpi}, in the literature, CNNs are used to extract features from videos and pictures and then passed to the SVM algorithm for categorization but cannot detect small-scale content in frames with different backgrounds. Hence, they propose to use the object detection algorithm YOLO \citep{Redmon16_CVPR} before the CNN, incrementing the detection accuracy. One of the most important studies on nude detection via video is the one by \citep{Lopes09_SIBGRAPI} where the authors use a Bag-of-Visual-Features, extracting from each frame-specific visual features (color, texture, shape, gradient) forming a visual vocabulary for clustering algorithms such as k-means. The study presented in \citep{Sanchez19_DI} is one of the first ones highlighting the stress forensics practitioners are exposed to while analyzing child abuse images and introduces the use of AI for the detection, processing evidence faster, with reduced false positives and content filtering. The study suggests more widespread training in data science and AI for forensic practitioners, the development of more effective filtering techniques, and the creation of standardized workflows to minimize the exposure of investigators to explicit material. In fact, NuDetective \citep{Castro10_IEEE} is one of the first tools developed several years ago to assist forensic examiners in identifying child pornography files but without using AI algorithms. NuDetective recognizes skin regions and assesses the likelihood of nudity according to the percentage of skin tones in image pixels and a filename association according to a specific database. Another interesting tool released on GitHub is the one developed by \footnote{\url{https://github.com/notAI-tech/NudeNet}} 
which recognizes and classifies the specific nude part, also giving a confidence level of the detected part by focusing on the context and meaning of the image and detecting explicit or inappropriate content by identifying body parts and their positions rather than just the amount of exposed skin. We used this tool, even if not forensic one, to check the detection of our dataset and compare the detection with the forensic tools. More works have been published on nudity detection without AI techniques, with simple pattern recognition based on pixel analysis or heuristic signature, such as \citep{Santos12_ISSPA, Garcia18_IEEE, Adnan16_Springer}. On the contrary, recent works have been published using AI for nudity detection, such as \citep{Moreira18_IJCNN} using different ML algorithms. Other publications have focused on revenge porn and sexting such as \citep{Mohanty19_MIPR} that developed a face recognition algorithm to look for the queried face in different social networks to detect if the person suffered revenge porn. The work published by \citep{Tariq19_Springer} identifies the main gaps in the research on the prevention of adolescent sexting behaviors, including the lack of pre-capture detection to prevent explicit content from being created in the first place and mobile platform-specific solutions and user-centered designs to consider the real-world behavior of adolescents. Different works have been published on \textit{sexual chat} predators detection based on emotions \citep{Bogdanova12_ACL} or other patterns \citep{Ebrahimi16_ISEIST, Ngejane21_FSIDI, Islam20_AsiaCCS}.

Based on the analysis of the current literature, we have seen that at the moment of writing this paper no current work analyses the tools commercially available now, even if different works addressed the topic.

\subsection{Previous Datasets}
\label{sec:sota:subsec:dataset}
In this section, we present the public datasets on the tests of our interest and explain whether these could be employed in our experimental analysis.
To test the robustness of the tools, we have to analyze their performance by loading different datasets in the evidence to be acquired. One of the tests we made is measuring the tools' behavior under deepfake images, and the most popular dataset is \citep{Yuezun20_cvpr}. This dataset contains images of popular actors and actresses taken from 590 popular YouTube videos and 5639 corresponding deepfake, with different ages, ethnic groups, and genders. The dataset has been labeled with a number corresponding to the real person and another one corresponding to the deepfake. Unfortunately, they did not release any matching between the name and the label number. For this reason we could not have used it in our experiments, instead of looking for the matching by hand, we preferred to create a new small dataset as explained in Section~\ref{sec:experiments:subsec:dataset}.

Another set of experiments are the ones related to pornography and child pornography. To the best of our knowledge, most of the tools online, even those already cited in Section~\ref{sec:sota:subsec:nude}, did not release any public dataset but only their methodology. The most popular one is the one released by \citep{Moreira16_FSI}, but looking online, it seems no longer available, and we did not receive any answer on time to conduct the experiments. Hence, as detailed in Section~\ref{sec:experiments:subsec:dataset}, we used almost the same methodology for our dataset creation. Instead, we found available for downloading the dataset \footnote{\url{https://universe.roboflow.com/tiem-
yhrf6/porn lab/dataset/1}} 
made up of about 12356 images, but all of them have a box highlighting the nude area. We claim that the box area could influence the pre-trained AI algorithm as the box is widely used in training. Hence, we decided not to use them to influence the black-box algorithm detection in the forensic tool. By analyzing some published papers and their bibliography, and with some free research on the Internet, we also found the dataset in \footnote{\url{https://huggingface.co/datasets/deepghs/nsfw detect/}} 
but we could not find any related publication concerning their methodology. As we are not sure how they scientifically made the dataset, we opted again to make it ourselves (in fact, we obtained some of the same or similar pictures, sometimes even identical, as described later in Section~\ref{sec:experiments:subsec:dataset}). Regarding child pornography, as everything is related to underaged people that must be protected, we could only find the dataset methodology acquisition and detection explained in \citep{macedo2018_sibgrapi} but no public data is available, also because developed with the collaboration of the Brazilian Police.

Regarding the weapons and drugs chats, we could not find any text dataset, but we found some public datasets on weapons pictures, while nothing for drugs. We decided to create the chats ourselves and send the found pictures on the chats. For the weapons, most datasets are about guns, and we used the following ones \citep{Delong20_arxiv} \footnote{\url{https://dasci.es/transferencia/open-data/24705/}} \footnote{\url{https://www.kaggle.com/datasets/snehilsanyal/weapon-
detection-test/data}}
\section{Experimental Analysis}
\label{sec:experiments}
The main goal of our work is to evaluate how AI-based DF tools behave when tasked to \emph{(i)} \textit{recognize nudity} 
in non-standard scenarios, \eg images portraying people with different skin tones, uncommon sexual acts, or unreal/fake but still recognizable bodies (i.e, mangas or dresses with nude drawings); \emph{(ii)} understand if we can \textit{bypass chat topic detection}; \emph{(iii)} study how the tool behaves in the case of \textit{deepfakes or images portraying people that underwent cosmetic or gender affirming surgery}. Due to the lack of suitable public datasets (as detailed in Section~\ref{sec:sota:subsec:dataset}), we created a dataset ex-novo, presented in Section~\ref{sec:experiments:subsec:dataset}. The dataset is balanced among categories. We strived to avoid introducing biases in the dataset (\eg skin tones, gender, etc). 
For reproducibility purposes, we performed multiple tests on different machines, ensuring the stability of the results. We also tested the results' case dependency by creating with the same tool multiple cases holding the same evidence, and checking the classification stability, \ie if pictures in the evidence are assigned the same label in all cases.

\subsection{Dataset Creation}
\label{sec:experiments:subsec:dataset}
To start our analysis, we looked for public nude image datasets and popular people deepfakes labeled with the whole name of the portrayed person. We could not find any public and ready-to-use dataset of images on the legal web about nudity. Of course, there are many deepfake face datasets due to the huge number of published works on the topic, but we could not use them because there is no list of the recognized popular people. For the previous reasons, we made all the datasets by ourselves. We collected $50$ images from Google with the following queries for fake nudes: designed nude t-shirt, naked t-shirt model, fake naked dress, naked clothes. For the other nudity questions, we looked at different pornography videos and saved some frames for a total of $50$ pictures for each category, including real nudity (with different skin tones and genders), teenagers, and unreal nudity (manga video). We had a total of $200$ pictures in the nude dataset.

Concerning the detection of popular people, we selected the actors Brad Pitt and Angelina Jolie (as they are one of the most popular couples of actors according to Google's statistics), collecting $50$ pictures of them and their deepfakes to understand if the name selection from the tool can recognize the deepfake as the real person. We also collected pictures of their child, Shiloh Jolie Pitt, to test for facial affinities. This last experiment was done to understand if the tool can confuse parents and children. Instead, to create the surgical dataset (persons who used some filler on the face like lips, cheekbone, nose, etc.) and the gender-affirming surgery dataset, we selected manually popular persons who were stated to belong to such categories from multiple public sources. Images constituting these datasets were gathered manually from Google searches and YouTube videos. This task could not be automated, since images portraying the same person belong to different categories, depending on when they were taken (\ie before or after surgery). Both the cosmetic and the gender-affirming surgery  datasets contain $50$ pictures, portraying $25$ people before and after surgery. We produced a total of $200$ pictures for face recognition.

Last but not least, we made $3$ different chats about drugs, weapons, and sex using Facebook Messenger in the Android app version, each containing more than $100$ messages. We employed the Android app in order to leverage the Magnet toolchain, where Magnet Acquire is used to dump the phone memory, which is subsequently given in input to Magnet Examine to search and analyze chat messages.
For the chat about drugs only $10$ messages have been sent with explicit content or using slang synonyms and $10$ pictures (including $2$ GIFs); the chat about weapons contains $25$ pictures taken from the public datasets as described in Section~\ref{sec:sota:subsec:dataset} and $10$ messages; about sex $50$ messages have been sent and $20$ pictures (with also $2$ pictures and $10$ messages about violence). We chose Facebook Messenger because it requires only an e-mail address for subscription; it is particularly suitable for creating fake digital personas (\eg, for scamming purposes).
In the chats, we sent different messages, also on different days, about the topic concerning the chat and used different language registers (\ie calling the drug by its common name or using different names such as snow, white powder, flour, talc, etc.) to understand which messages and multimedia would be detected.

In summary, we created the datasets for the following topics:
\begin{itemize}
    \item $200$ images for nude detection, including: real porn images, actors with teenager appearance, dresses with drawn sensitive body parts, mangas depicting sexual activities;
    \item $200$ images for face recognition: popular real actors detection and their deepfakes, with also some pictures of their children, popular people who underwent cosmetic or gender-affirming surgery;
    \item $3$ chats, one for each topic sending formal and slang messages and pictures about sex, drugs and weapons. 
    
\end{itemize}

\subsection{Digital Forensics Tool}
After building the dataset, we tested the DF tools described in Section~\ref{sec:foren_tools:subsec:tools_ai}. We performed different experiments on a per-tool basis since such tools are able to detect different types of evidence. Thus, considering that no evidence category is detected by both tools, we could not perform a direct comparison of the detection accuracy among the tools.

\subsubsection{Magnet AI}
\label{sec:experiments:subsec:magnet}
As described in Section~\ref{sec:foren_tools:subsec:tools_ai}, Magnet Axiom Examine uses Magnet AI to detect specific messages on chats or multimedia on the acquired evidence and the tools come together. The topics Magnet AI can recognize are the following: animals, bedrooms, buildings (exterior), child abuse, documents (card/ID), documents (paper), drones/UAVs, drugs, handwriting, hate symbols, human faces, human hands, icons, invoices/receipts, license plates, militants, money, nudity, screen captures, tattoos, vehicles (cars, trucks, vans, buses), weapons; whereas for chat only grooming/luring and sex-related. 
For this reason, we asked the following questions, which will become the experimental line:
\begin{itemize}
    \item What happens when we load interracial nude pictures? Does it recognize them properly, or does it have some bias based on the skin tone?
    \item What happens if we load pictures of dresses with a painted nude?
    \item Does the tool recognize unreal nude pictures such as those extracted from pornographic comics videos?
    \item Does it recognize the teenage actors (which are all people $18$+ years old according to the disclaimers) in the pornography videos? 
    \item What happens for the chat? Does it recognize the pictures sent, or can it deduce the topic even if the users talk using metaphors that a human can clearly understand given the chat context?
\end{itemize}
\subsubsection{Excire Photo AI}
\label{sec:experiments:subsec:excire}
As described in Section~\ref{sec:foren_tools:subsec:tools_ai}, the tool X-Ways Forensics uses Excire Photo AI to detect common objects and faces of known popular people and can also be downloaded separately with a trial version. The list of known objects is publicly released\footnote{\url{https://www.x-ways.net/Excire\_Detected\_Objects.txt}}, while the list of recognized popular people is not disclosed. With Excire Photo AI, we can search in the loaded dataset for a specific popular person name, and it will display the results for that name if present in the training set (hence we can do some inquiry to try to understand the training dataset even if should be very large) and also select a face and look for similar faces in the loaded dataset. For this reason, starting from the homemade dataset, we asked the following questions:
\begin{itemize}
    \item Can it recognize a popular actor real face?
    \item What happens when we load a deepfake face of the actor? Is it labeled as the fake one or the original one?
    \item What happens if I type the name of a popular actress but I have in the dataset even the pictures of the biological children (for facial similarities)?
    \item Is the tool powerful enough to recognize a popular person before and after facial surgery? 
    \item How does facial recognition behave if a previously observed person undergoes gender affirming surgery? 
\end{itemize}
\section{Experimental Procedure and Results}
\label{sec:results}
In this section, we will detail the experimental procedure and present the results of the different experiments with the two analyzed tools. Overall, we can say that the tools can recognize most of the subjects well, but sometimes, or in some situations, they are misled by the content, resulting in different false positives.

\subsection{Magnet AI}
\begin{figure}[h]
    \centering
    \includegraphics[width=0.35\textwidth]{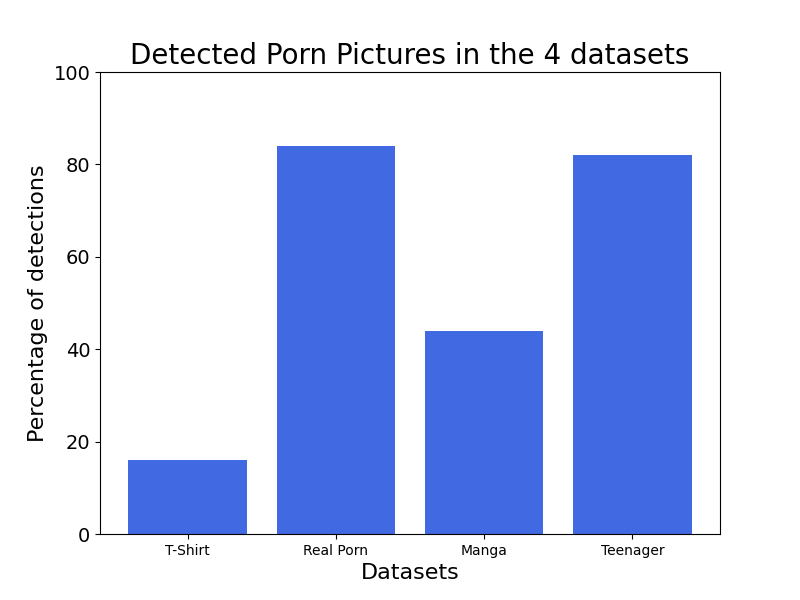}
    \caption{The bar chart shows the percentage of nude images detected as nude by Magnet AI tool for each of the examined $4$ categories: t-shirt, real porn, manga and teenager, each one containing $50$ pictures.}
\label{sec:results:fig:nudes}
\end{figure}

We made $4$ different acquisitions, one for each dataset category, of a USB pendrive containing the $50$ pictures belonging to each category, and used the Magnet "Analyse Pictures with AI" to understand how many pictures would be detected as nudity. Results are shown in Figure~\ref{sec:results:fig:nudes} where a perfect classifier should have a $100$\% detection rate for each category.

\begin{figure}[h]
    \centering
    \includegraphics[width=0.35\textwidth]{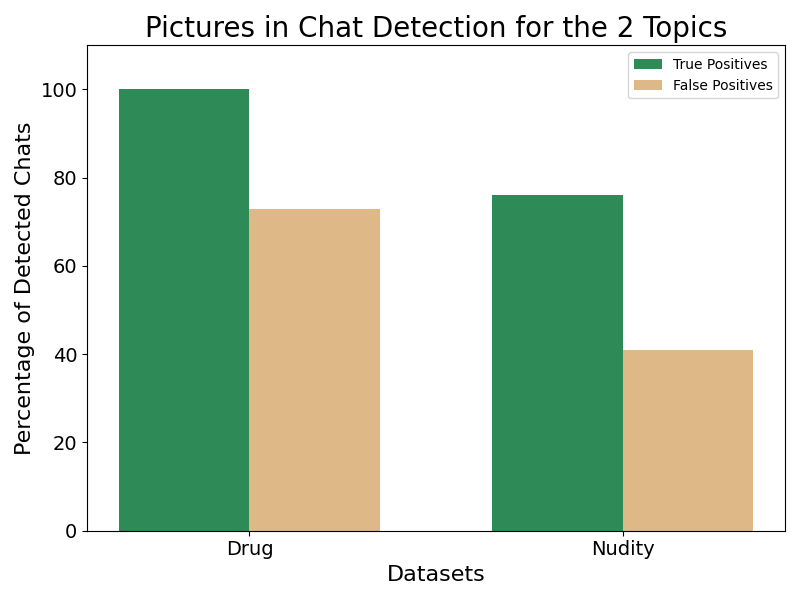}
    \caption{The bar chart shows the percentage accuracy for the chat tests made with the Magnet AI analysis. The dataset is made of drug and nude pictures. For each analyzed dataset, the left green bar is the samples detected correctly (true positives), while the yellow right bar represents the other images labeled as those categories (false positives).}
    \label{sec:results:fig:chats}
\end{figure}

We also tested Magnet's feature "Analyse Chats with AI". Unfortunately, there are no publicly available datasets (as reported in Section~\ref{sec:sota:subsec:dataset}) of chats about prostitution, drugs and weapons trafficking. For this reason, we opted to write 3 different chats pertaining these topics on the Facebook Messenger Android app. We wrote one chat for each topic and sent in each chat different formal and slang messages (\eg calling the drug as snow, white powder, flour, etc.), along with related pictures, in order to comprehend the ability of the tool to detect the topic. 
The tool only labeled sex/nudity and drugs classes in the sent pictures. Figure~\ref{sec:results:fig:chats} only shows results of this search. Moreover, no results are displayed regarding the chat categorization for drug and sex/nudity since the tool "Analyze chats with AI" feature was not able to flag any of our chat messages, explicit messages or metaphoric ones. 

\subsection{Excire Photo AI}
With this tool, used by X-Ways Forensics but that can also be found separately and used with a trial-free license, we made all the tests regarding face recognition. With popular actors, we mean the set of experiments where a popular name of an actor or actress has been typed with the prompt-AI feature. In this test, we also included some deepfakes (false positive pictures) of the same people. We expect that some will be recognized as the same real and original person because deepfake detection is still an open research topic. Parents experiments are those made with the Pitt-Jolie family and trying to recognize the children Shiloh Jolie-Pitt by typing the name or by selecting the face and looking for similar pictures, trying to understand if the parents' similarities are denoted by the tool. Instead, with the last two experiments, we wanted to understand if, after a cosmetic or gender affirming surgery, the tool can recognize the person both by typing the name in the command prompt and by selecting the face and looking for similar pictures. We expect that the tool is trained to still recognize them, at least by typing the name.

\begin{figure}[h]
    \centering
    \includegraphics[width=0.35\textwidth]{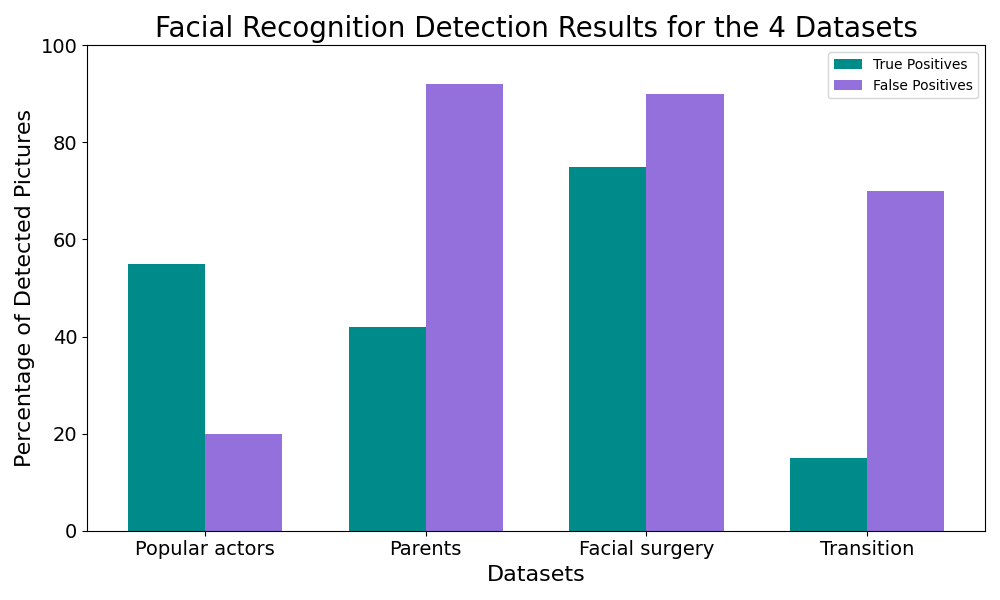}
    \caption{The bar chart shows the accuracy of the Excire Photo AI picture detection algorithm, in terms of percentage of correctly detected faces contained in the datasets described in Section~\ref{sec:experiments:subsec:dataset} (true positives, green bar on the left) and the percentage of different pictures erroneously classified as portraying the queried person (false positives, violet bar on the right).}
    \label{sec:results:fig:face}
\end{figure}

\section{Discussion}
\label{sec:discussion}
In this section, we discuss the results of our experiments on the two DF tools. According to these results, we also answer to the Research Questions provided in Section~\ref{sec:intro}.

\subsection{Magnet AI}
\label{sec:discussion:subsec:magnet}
\begin{figure}[h]
    \centering

    \begin{subfigure}[b]{0.5\textwidth}
        \centering
        \includegraphics[width=\textwidth]{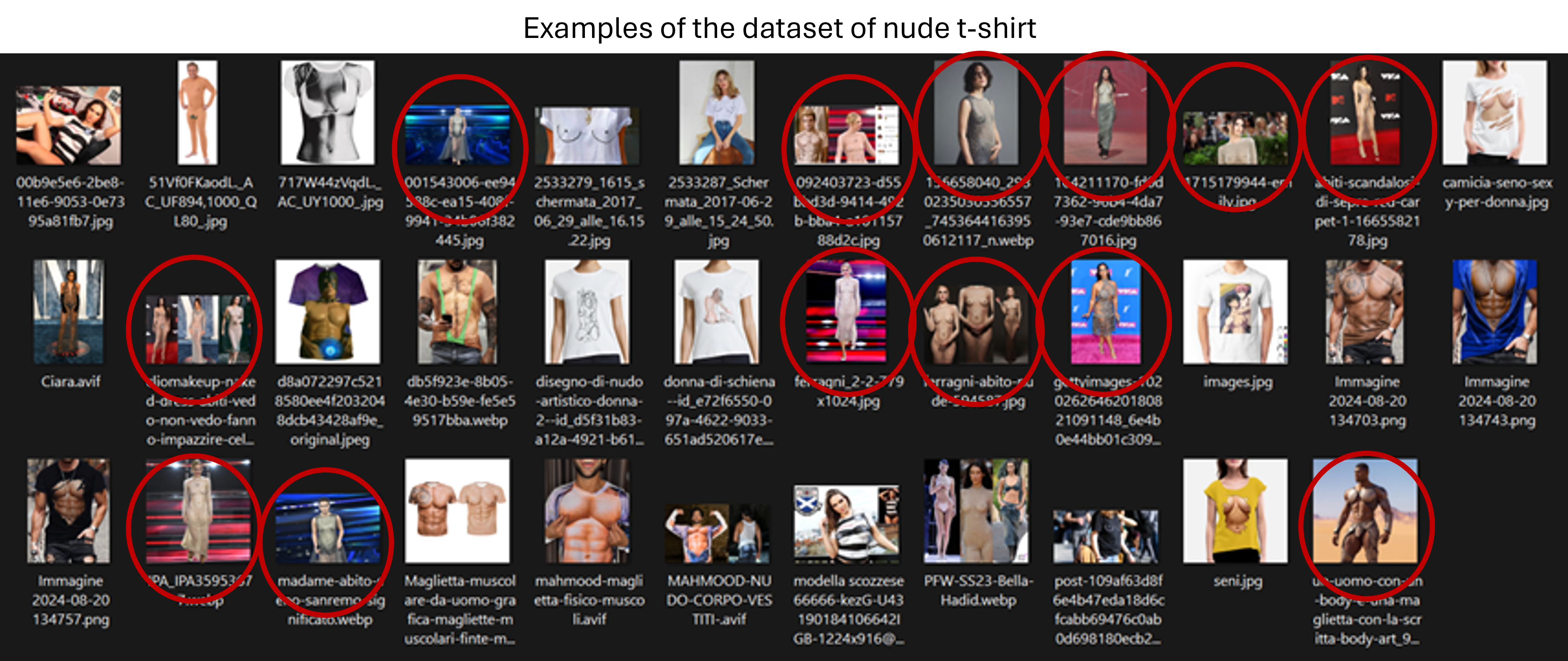}
        \caption{Portion of the dataset used for nude t-shirts and dresses. In the red circle, the very real body is painted on the dresses.}
        \label{sec:discussion:fig:dataset-tshirt}
    \end{subfigure} 
    \hfill
    \begin{subfigure}[b]{0.4\textwidth}
        \centering
        \includegraphics[width=\textwidth]{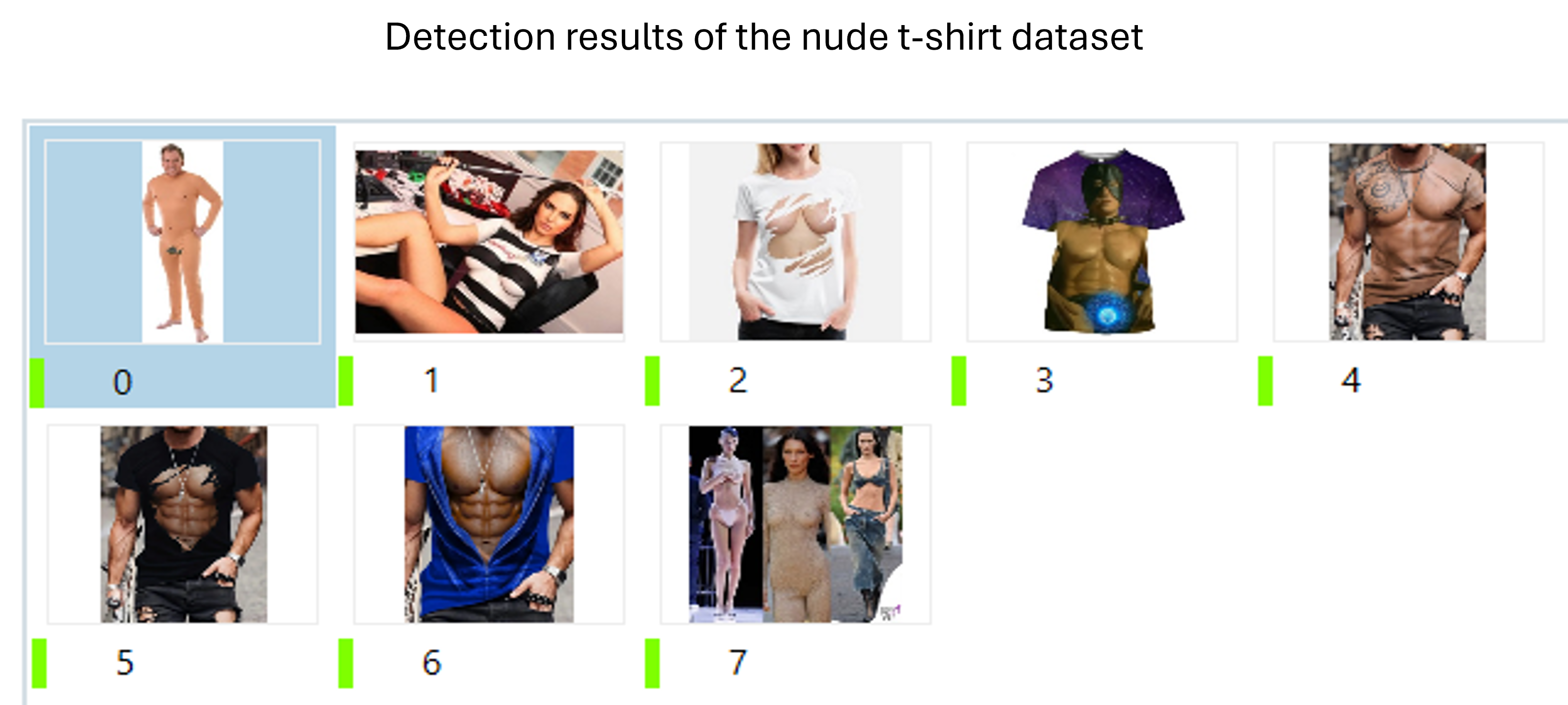}
        \caption{Pictures of dresses with painted nudes detected as nude.}
        \label{sec:discussion:fig:detected-tshirt}
    \end{subfigure}
\caption{Dataset of nude t-shirts (top) and the detected results (bottom) by the Magnet AI tool for nude detection experiments on painted dresses.}
\label{sec:discussion:fig:tshirt-detection}
\end{figure}

From the results presented in Figure~\ref{sec:results:fig:nudes}, we can say that the tool correctly recognizes real nude pictures (real porno and teenagers) even with different skin tones or gender diversity, but not all of them are recognized as nude. The undetected pictures are not so different from the detected ones in terms of content and dimension/light/contrast, so it is unclear why have not been marked as nude. The same happens for the manga category because humans can correctly identify the pictures as nude, but maybe some features in the AI algorithm cannot mark them as nude. Regarding the t-shirt results, we did not expect such a high detection rate after looking at the pictures labeled as nude. In the dataset, we put some dresses with a real painted woman's body that we thought would be recognized as nude $100\%$ (the red circled ones in Figure~\ref{sec:discussion:fig:dataset-tshirt}) and also some pictures that a human can easily recognize as a t-shirt or a dress with some paintings (such as image $3$ in the Figure~\ref{sec:discussion:fig:detected-tshirt}). We loaded the pictures on NudeNet, and all the pictures have been correctly identified because the latter recognizes and classifies specific parts of the body rather than calculating the percentage of exposed skin (we suppose that Magnet AI is instead based on this last detection technique).

\begin{figure}[h]
    \centering
    \includegraphics[width=0.4\textwidth]{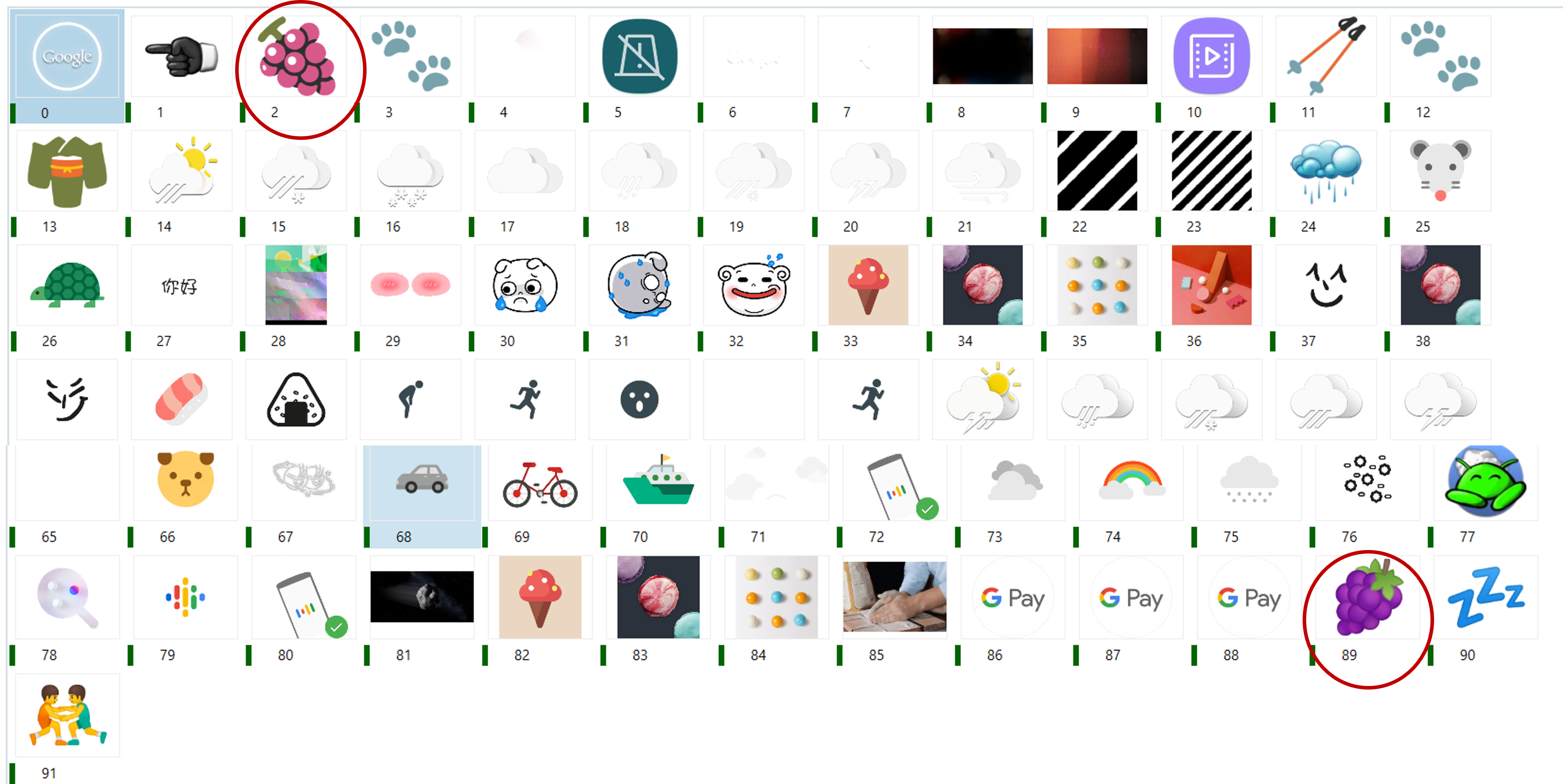}
    \caption{The image shows the pictures detected as drugs but that clearly are not drugs (false positive rates) such as the grape (red circled) for the drug detection in chat done by Magnet AI tool.}
    \label{sec:discussion:fig:drug}
\end{figure}

Regarding the pictures sent in the chat (results shown in Figure~\ref{sec:results:fig:chats}), the false positive rates include also other default pictures found in the Android device memory dump. Even if the dataset has been made only of pictures belonging to those categories, when dumping the memory of the Android device using Magnet Acquire, the dump contained other pictures stored in the device, such as stickers or default pictures (shown in Figure~\ref{sec:discussion:fig:drug}) that have been categorized as drugs even if it is humanly clear they are not drugs nor nude pictures.

\subsection{Excire Photo AI}
\begin{figure}[h]
    \centering
    \includegraphics[width=0.3\textwidth]{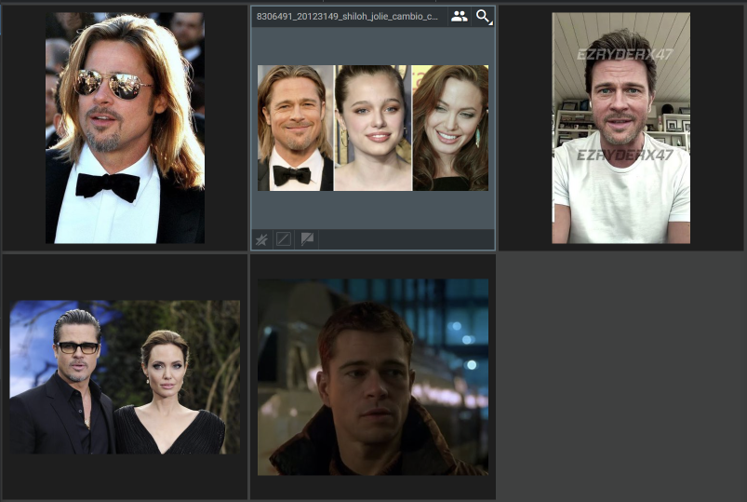}
    \caption{Obtained results for querying Brad Pitt in the Excire Photo AI tool.}
    \label{sec:discussion:fig:brad}
\end{figure}

From the graph in Figure~\ref{sec:results:fig:face} we clearly see that the detection rate of real pictures is not that high and on average it is higher for false positives detection. This means that the tool performs badly on average, recognizing as the same person pictures not related with the same person, both by typing the name in the command prompt and by finding similarities on the face. For example, as shown in Figure~\ref{sec:discussion:fig:brad} looking for Brad Pitt pictures, between $20$ pictures the tool only recognizes $3$ real of them and $2$ deepfakes. Other $3$ deepfakes in the dataset have not been recognized similar or belonging to "Brad Pitt" class. This could depend on the deepfake quality and is a good result. 

\begin{figure}[h]
    \centering
    \includegraphics[width=0.35\textwidth]{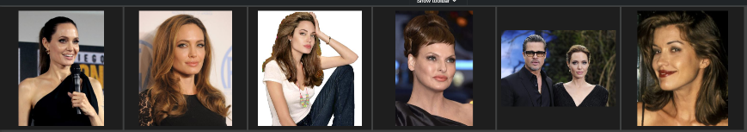}
    \caption{Results for querying pictures of Angelina Jolie in the Excire Photo AI tool.}
    \label{sec:discussion:fig:angelina}
\end{figure}

\begin{figure}[h]
    \centering
    \begin{subfigure}[b]{0.4\textwidth}
    \centering
    \includegraphics[width=\textwidth]{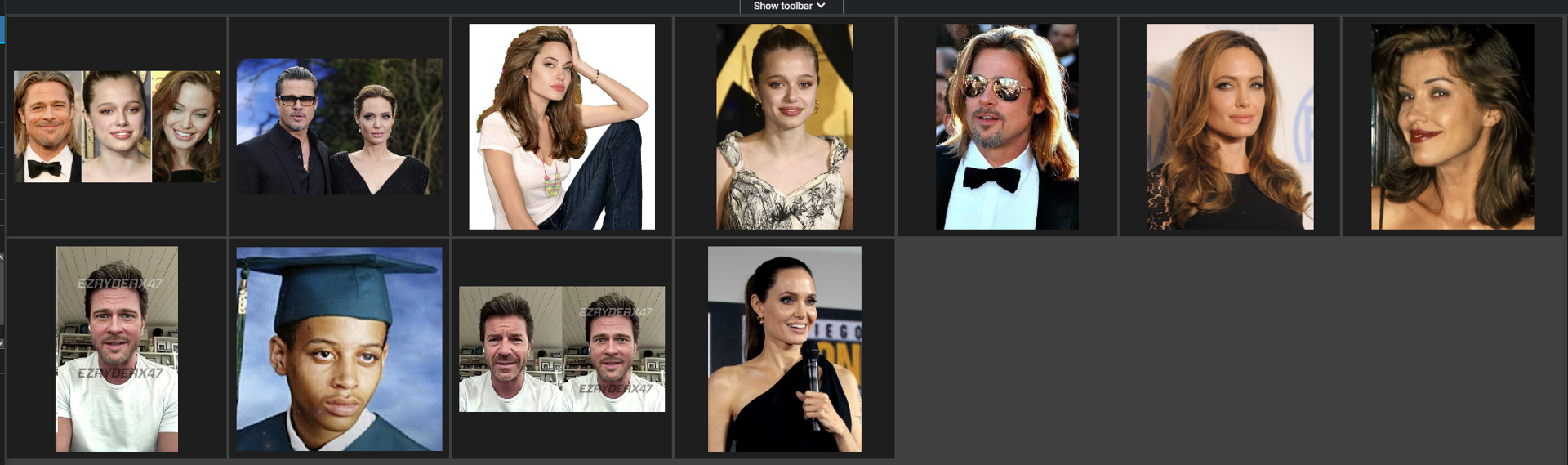}
    \caption{Results when querying "Shiloh Jolie-Pitt" to check parents' similarities.}
    \label{sec:discussion:fig:shilohjoliepitt}
        
    \end{subfigure}
    \hfill
    \begin{subfigure}[b]{0.4\textwidth}
    \centering
    \includegraphics[width=\textwidth]{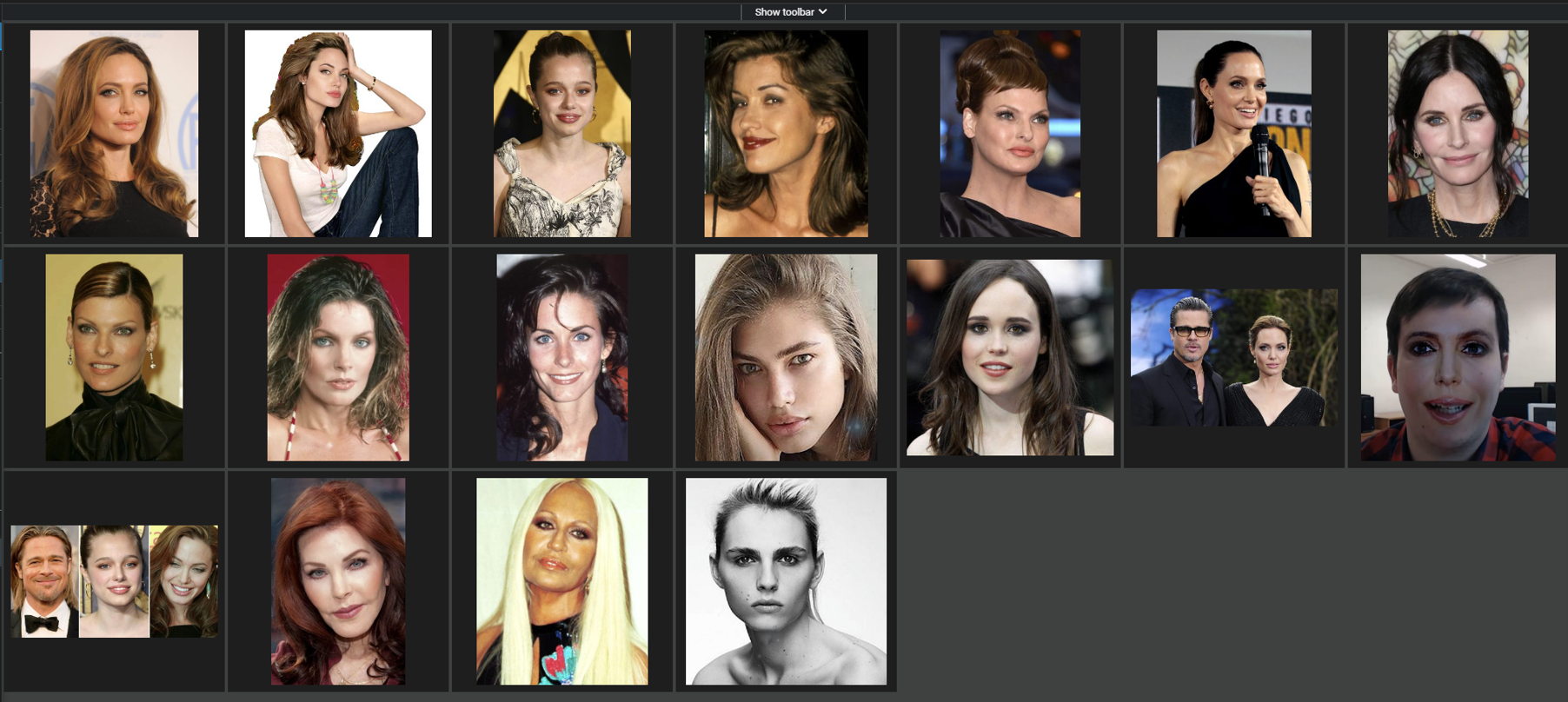}
    \caption{Results when querying Jolie to check parents' similarities.}
    \label{sec:discussion:fig:jolieshiloh}
        
    \end{subfigure}
    \hfill
    \begin{subfigure}[b]{0.35\textwidth}
    \centering
    \includegraphics[width=0.8\textwidth]{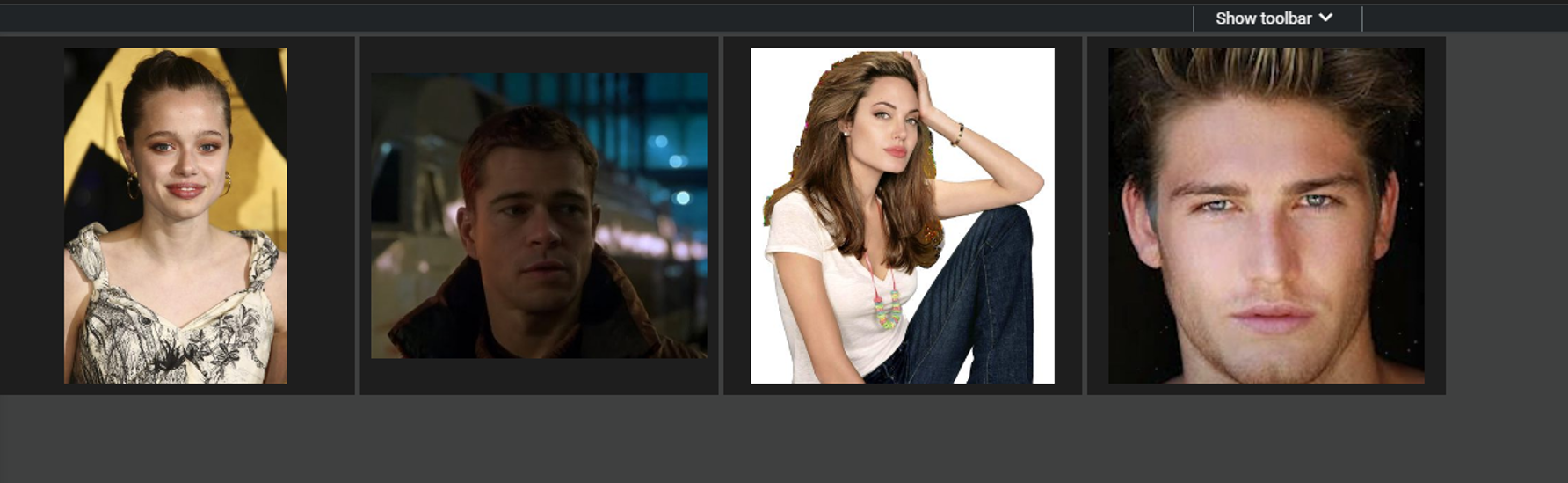}
    \caption{Results when querying Pitt to check parents' similarities.}
    \label{sec:discussion:fig:pittshiloh}
    \end{subfigure}
    
    \caption{Overall results about parents' surname querying in the Excire Photo AI tool.}
    \label{sec:discussion:fig:parents}
\end{figure}

\begin{figure}[h]
    \centering
    \includegraphics[width=0.3\textwidth]{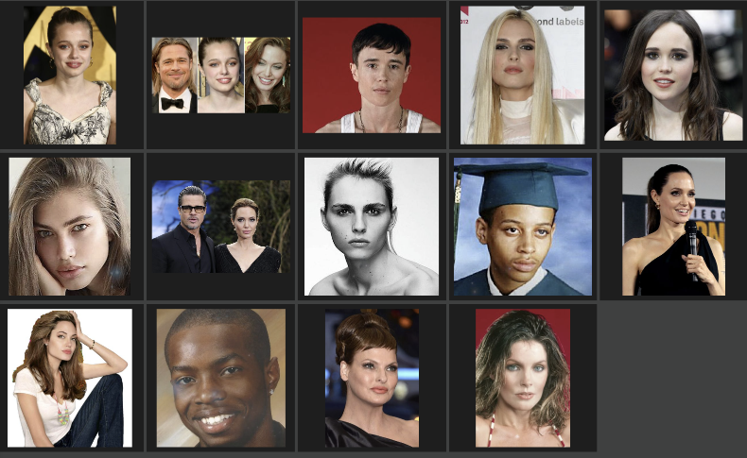}
    \caption{Results by selecting Shiloh's face and looking for face similarities, a feature of Excire Photo AI tool.}
    \label{sec:discussion:fig:shiloh}
\end{figure}

When looking for parent similarities, we first tried to type the entire name "Shiloh Jolie-Pitt" and found the results shown in Figure~\ref{sec:discussion:fig:shilohjoliepitt}. As we can see, it detected a deepfake of Brad Pitt and two pictures not related with none of them. This misclassification has been found also when looking for "Jolie" in Figure~\ref{sec:discussion:fig:jolieshiloh} (with also a deepfake not detected for clear "Angelina Jolie" results, Figure~\ref{sec:discussion:fig:angelina}) and "Pitt" results, Figure~\ref{sec:discussion:fig:pittshiloh}. With this tool it is very common to have other people's pictures when typing the name. For this reason we selected the face of Shiloh and looked for similar faces. The results displayed in Figure~\ref{sec:discussion:fig:shiloh} show other people clearly not related or similar to the queried face. We also claim that the displayed pictures are sorted according to some similarity algorithm, but it has not been declared by the tool's proprietaries.

From all the results displayed regarding the Pitt-Jolie family, we can say that sometimes the images are random because a deepfake detected as similar twice for Brad Pitt (the man with the white t-shirt) has not been detected in the Pitt query (Figure~\ref{sec:discussion:fig:pittshiloh}).

\begin{figure}[h]
    \centering
    \begin{subfigure}[b]{0.3\textwidth}
    \centering
    \includegraphics[width=0.8\textwidth]{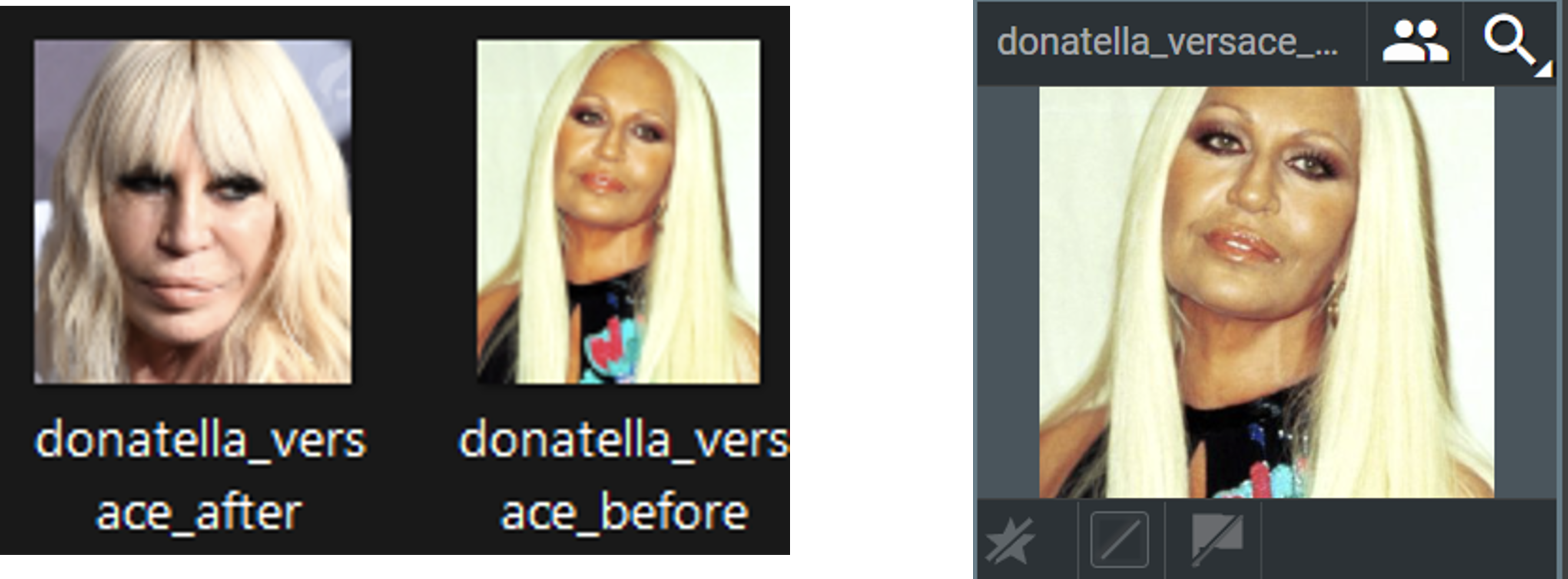}
    \caption{Detection results for Donatella Versace before and after surgery, only before detected.}
    \label{sec:discussion:fig:versace}
        
    \end{subfigure}
    \hfill
    \begin{subfigure}[b]{0.4\textwidth}
    \centering
    \includegraphics[width=\textwidth]{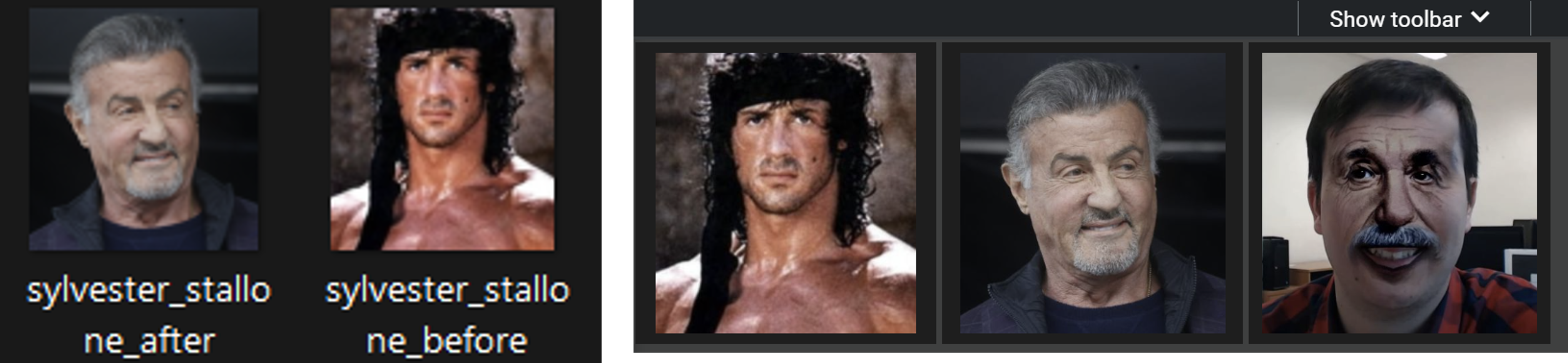}
    \caption{Detection results for Sylvester Stallone, detected before and after surgery and a false positive that is a bad deepfake of Albert Einstein.}
    \label{sec:discussion:fig:stallone}
        
    \end{subfigure}
    \hfill
    \begin{subfigure}[b]{0.4\textwidth}
    \centering
    \includegraphics[width=\textwidth]{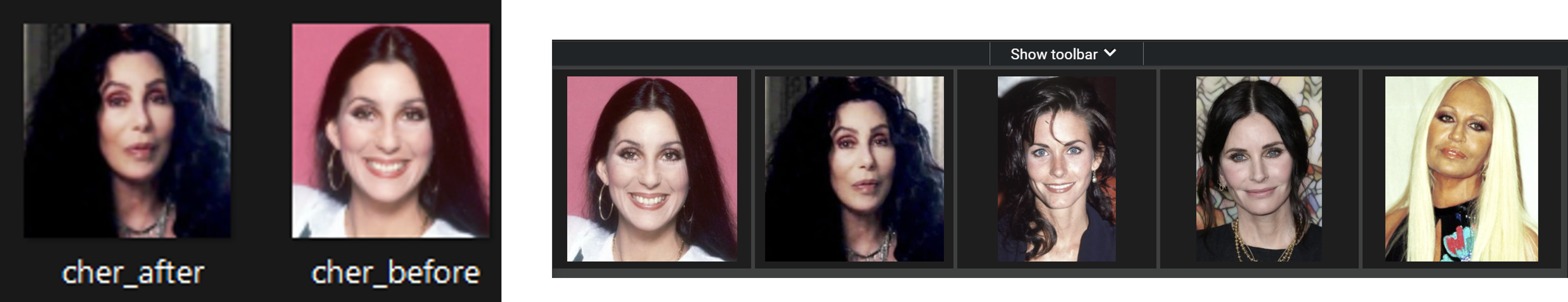}
    \caption{Detection results for Cher, detected before and after surgery but also Donatella Versace before surgery and Courtney Cox before and after ($3$ false positives).}
    \label{sec:discussion:fig:cher}
        
    \end{subfigure}
    \hfill
    \begin{subfigure}[b]{0.4\textwidth}
    \centering
    \includegraphics[width=\textwidth]{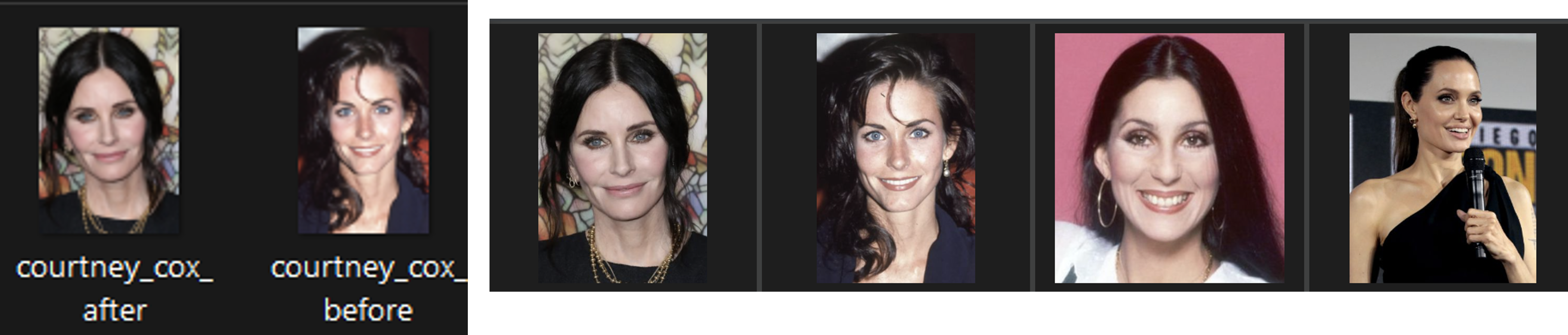}
    \caption{Detection results of Courtney Cox, before and after surgery but also $2$ false positives: Cher before, Angelina Jolie.}
    \label{sec:discussion:fig:ccox}
        
    \end{subfigure}
    \caption{Detection results for the cosmetic surgery dataset analysed by Excire Photo AI: on the left the dataset and on the right the detection results.}
    \label{sec:discussion:fig:surgery}
\end{figure}

About the facial surgery tests, as shown in the bar chart of Figure~\ref{sec:results:fig:face}, it is the test with the highest results for correct detections. Despite this, it also has a high rate of false positives. In fact, as shown in Figure~\ref{sec:discussion:fig:surgery}, for some people like Donatella Versace in Figure~\ref{sec:discussion:fig:versace}, the tool cannot recognize the same person by name in the picture after facial surgery. While for other people like Sylvester Stallone in Figure~\ref{sec:discussion:fig:stallone}, Cher in Figure~\ref{sec:discussion:fig:cher}, and Courtney Cox in Figure~\ref{sec:discussion:fig:ccox}, it can detect the pictures before and after surgery, but also other people with no evident resemblance to the target person.

About trans people, the tool is not able to find the persons by name. Maybe they have not been used in the algorithm training dataset. Thus, we used the facial similarities feature and selected the face for each of the $25$ people in the dataset before and after the transition. In very few queries, the tool recognized the pictures of the same person but most of the time recognized other people. To respect their privacy and dignity, we do not show neither the detection results nor the related pictures.

\subsection{Addressing the Research Questions}
Based on the results presented in Section~\ref{sec:experiments}, we provide our answers to the research questions presented in Section~\ref{sec:intro}. 

\emph{RQ1.} First, these tools are not robust enough in the detection as shown by the high number of misclassifications (\eg common objects are detected as suspicious, like the grape results in Figure~\ref{sec:discussion:fig:drug}). According to the bar charts in Figure~\ref{sec:results:fig:chats} and Figure~\ref{sec:results:fig:face} we have many false positives, \ie pictures belonging to other classes that are recognized as the target one. Furthermore, the true positive detection rate (number of samples belonging to a class and correctly classified) depends on the target class (\eg it can detect all drug pictures, instead has some errors for nudity and face recognition).

\emph{RQ2.} Concerning face recognition we report that some deepfakes are recognized as the real person, as show in Figure~\ref{sec:discussion:fig:brad}. We expected this result since deepfake detection is still an open research work. Nevertheless, the forensics analyst cannot trust the labels assigned automatically by the tool, and must still check them manually. Moreover, the results of our experiments on parents and children pictures show that similarities among different subjects in the analyzed photo corpus may lead to misclassifications. It is even more worrisome that sometimes, when selecting the face of a certain person and looking for similar faces in the whole evidence, the tool recognizes people that have nothing in common (Figure~\ref{sec:discussion:fig:angelina}, Figure~\ref{sec:discussion:fig:parents}, Figure~\ref{sec:discussion:fig:shiloh}, Figure~\ref{sec:discussion:fig:surgery}). 

\emph{RQ3.} Considering the experimental results, especially false positives, we claim that more advanced techniques can be developed to hide secret data from being detected by AI-based forensics tools in an anti-forensics techniques scenario.

\section{Conclusion and Future Work}
\label{sec:future_work}

In this work, we analysed the robustness of Artificial Intelligence black box algorithms used in Digital Forensics to understand how such algorithms may be bypassed (adversarial attacks scenarios) by cybercriminals and expert users to avoid detection of their sensitive data in legal proceedings. According to the reached results we can say that AI is far from ready to substitute humans in the entire analysis task. Indeed, human experts must still supervision such algorithms, full checking their results. Nevertheless, such algorithms can be indeed useful to provide a starting point for human analysis, and in general to ease the burden of human expert.

Starting from the results presented in this work, we briefly report on how we plan to extend our research. First, we plan on improving our testing with a larger dataset and a novel methodology to automatically create it. Then, we will expand the testing scenario by thinking about different tests we may perform to bypass the algorithm. Regarding chat, we plan to use Large Language Model (LLM) algorithms to generate an augmented dataset of chats according to specific topics and even generate nude pictures, of course respecting legal obligations. We will test the performances of the existing ones and also train specific LLMs on this topic. The dataset could also be improved by considering quality, dimensions, lights and other important factors for image detection.

The analyzed tools should improve the robustness to recognize all the pictures in the testing dataset, hence improving the training with more test cases as the ones highlighted in our work. For this reason, we propose to make a platform similar to the original algorithm, keeping it proprietary and black-boxed so that the community can test adversarial examples or other testing cases and the company can retrain the algorithm. Moreover, as highlighted by other works, we claim that using xAI can help the forensic analyst understand why the image has been categorized with a specific label instead of another one or why it has not been labeled. With xAI, we claim that we can also reduce the set of possible adversarial attacks because we know how and why the algorithm fails.

In general, we think that AI can better assist forensics analysts in their job by filtering specific contents without watching the entire evidence and knowing private data of the defendant not related to the prosecution object or without being shocked by watching strong pictures (i.e., child abuse, violence, homicide). We also think that AI can help in general in the analysis, being a sort of digital assistant so that the algorithm makes a first analysis and a first report and then helps the analyst suggest specific actions. In this way, the forensics job is lighter, but as the AI is trained on much more cases than the human analyst, we can combine the two knowledge. Of course, we have to pay attention to adversarial examples, which are also used for anti-forensics purposes, because criminals can also use this system to understand how to hide data.











\bibliographystyle{elsarticle-harv}
\bibliography{elsarticle-template-harv}

\end{document}